\documentclass[conference]{IEEEtran}
\IEEEoverridecommandlockouts
\usepackage{amsmath,epsfig}

\usepackage{cite}
\usepackage{amsfonts}
\usepackage{url}
\usepackage{algorithm}
\usepackage{algorithmic}
\usepackage{colortbl}
\usepackage[breaklinks, hidelinks]{hyperref}
\usepackage[nolist, nohyperlinks]{acronym}
\usepackage{bm}
\usepackage{balance}
\usepackage{comment}
\usepackage{orcidlink}
\usepackage{adjustbox}
\usepackage{pifont}
\usepackage{multirow}
\usepackage[caption=false,font=footnotesize,subrefformat=parens,labelformat=parens]{subfig}
\usepackage[normalem]{ulem}
\useunder{\uline}{\ul}{}

\begin{acronym}
\acro{AE}{Autoencoder}
\acro{AI}{Artificial Intelligence}
\acro{BCE}{Binary Cross-Entropy}
\acro{CNN}{Convolutional Neural Network}
\acro{CPS}{Cyber-Physical System}
\acro{DDoS}{Distributed Denial-of-Service}
\acro{DoS}{Denial-of-Service}
\acro{ELBO}{Evidence Lower Bound}
\acro{ELU}{Exponential Linear Unit}
\acro{FCN}{Fully Convolutional Network}
\acro{GAN}{Generative Adversarial Network}
\acro{GNN}{Graph Neural Network}
\acro{ICA}{Independent Component Analysis}
\acro{IDS}{Intrusion Detection System}
\acro{IoT}{Internet of Things}
\acro{IR}{Information Retrieval}
\acro{LR}{Logistic Regression}
\acro{LSTM}{Long Short-Term Memory}
\acro{GMAC}{Giga Multiply-Accumulate Operation}
\acro{MSA}{Multivariate Statistical Analysis}
\acro{RF}{Random Forest}
\acro{RNN}{Recurrent Neural Network}
\acro{SVM}{Support Vector Machines}
\acro{PCA}{Principal Component Analysis}
\acro{PFA}{Probability of False Alarm}
\acro{VAE}{Variational Autoencoder}
\end{acronym}

\definecolor{green_cool}{rgb}{0.0, 0.5, 0.0}
\definecolor{blue_cool}{rgb}{0.12, 0.32, 1}
\definecolor{red_cool}{rgb}{0.5, 0.0, 0.0}
\definecolor{Gray}{gray}{0.85}

\def\BibTeX{{\rm B\kern-.05em{\sc i\kern-.025em b}\kern-.08em
    T\kern-.1667em\lower.7ex\hbox{E}\kern-.125emX}}

\begin{document}

\title{Unsupervised Network Anomaly Detection with Autoencoders and Traffic Images
\thanks{The research presented in this paper was partially funded by the project “ISEEYOO: AI-based Network Anomaly Detection for CPS exploiting 2D data representation” within the University of Padova funding framework ``SEED research grants.''}
}


\author{\IEEEauthorblockN{Michael~Neri\IEEEauthorrefmark{1}~\orcidlink{0000-0002-6212-9139}, Sara~Baldoni\IEEEauthorrefmark{2}~\orcidlink{0000-0001-5642-3430}}
\IEEEauthorblockA{\IEEEauthorrefmark{1}\textit{Faculty of Inform. Tech. and Com. Sciences}, \textit{Tampere University}, Tampere, Finland, \href{mailto:michael.neri@tuni.fi}{michael.neri@tuni.fi}}
\IEEEauthorrefmark{2}\textit{Dept. of Information Engineering}, \textit{University of Padova}, Padua, Italy, \href{mailto:sara.baldoni@unipd.it}{sara.baldoni@unipd.it}}

% \author{\IEEEauthorblockN{Michael Neri}
% \IEEEauthorblockA{\textit{Faculty of Inform. Tech. and Com. Sciences} \\
% \textit{Tampere University}\\
% Tampere, Finland, \\
% michael.neri@tuni.fi}
% \and
% \IEEEauthorblockN{Sara Baldoni}
% \IEEEauthorblockA{\textit{Dept. of Information Engineering} \\
% \textit{University of Padova}\\
% Padua, Italy \\
% sara.baldoni@unipd.it}
% }


\maketitle

\begin{abstract}
Due to the recent increase in the number of connected devices, the need to promptly detect security issues is emerging. Moreover, the high number of communication flows creates the necessity of processing huge amounts of data. Furthermore, the connected devices are heterogeneous in nature, having different computational capacities. For this reason, in this work we propose an image-based representation of network traffic which allows to realize a compact summary of the current network conditions with 1-second time windows. The proposed representation highlights the presence of anomalies thus reducing the need for complex processing architectures. Finally, we present an unsupervised learning approach which effectively detects the presence of anomalies. The code and the dataset are available at \url{https://github.com/michaelneri/image-based-network-traffic-anomaly-detection}.
\end{abstract}


\section{Introduction}\label{sec:Intro}

The current diffusion of Internet-related technologies is leading to an unprecedented connectivity among heterogeneous devices with varied computational capabilities.
%Currently, the Internet and computer networks have achieved information interconnectivity, hastened information transmission, and transformed the manner in which data is conveyed. 
The extensive use of computer networks creates security risks that can lead to misconduct and substantial harm. These threats are dynamic and prone to evolve into unknown forms~\cite{Lin_2019_Cloud}. To properly react to this danger, a prompt detection of anomalous network behaviors is needed.
An \textit{anomalous} event can be defined as a network pattern that diverges from the expected \textit{normal} behavior~\cite{Chandola_2009_ACMComputing}. 
The design of anomaly detection techniques is challenging for several reasons. First, the wide diffusion of connected devices causes a relevant increase in the number of traffic flows, making the real-time detection of anomalies a demanding task. Moreover, due to the inherent disparity between the amount of normal and anomalous data flows, the adoption of supervised learning methods is hindered. Furthermore, these techniques often fail in accurately identifying unfamiliar abnormal behaviors~\cite{Cook_2020_IoT, Ahmad_2021_TETT}.  Consequently, the exploration of unsupervised learning techniques has emerged as a prominent direction for addressing anomaly detection within telecommunication networks. 
In the context of unsupervised techniques, a key task consists in modeling the normal state of a telecommunication network. To this end, different types of predictors such as traffic usage, protocols, and number of flows can be employed~\cite{Ahmad_2021_TETT, Ahmad_2022_ICSTW}. 
% Before PCA-based methods, there were Multivariate Statistical Analysis methods
%Traditional methods for detecting anomalies in network traffic rely on the \ac{MSA}~\cite{Dryden_JSTOR_1993} paradigm. More specifically, it deals with multiple observations of a phenomena that have some inherent interdependence to predict one or more outcomes. The authors in ~\cite{Guangzhi_ICCSA_2005} introduced an online multivariate analysis algorithm for proactive detection of network attacks, including viruses, worms, and \ac{DDoS} attacks. This method analyzes system resource behaviors and network protocols to identify deviations from normal operational states. Similarly, Bodenham \textit{et al.}~\cite{Bodenham_ICDMW_2013} developed a statistical anomaly detection methodology for continuous monitoring of network traffic, utilizing multivariate adaptive estimation to reduce the need for multiple control parameters. 
%This approach, validated on simulated and real NETFLOW data, focuses on monitoring traffic through different TCP ports with a single control parameter for practical ease of use by network analysts. 
% PCA BASED Network anomaly detection
Due to the high number of predictors, dimensionality reduction techniques, such as \ac{PCA}~\cite{Mackiewicz_CompGeo_1993}, have been employed for analyzing network traffic~\cite{Callegari_ICC_2011, Camacho_CCW_2014, Nguyen_2023_INFOCOM}. 
\begin{figure*}[ht!]
     \centering
     \subfloat[Normal]{\includegraphics[width=0.27\linewidth]{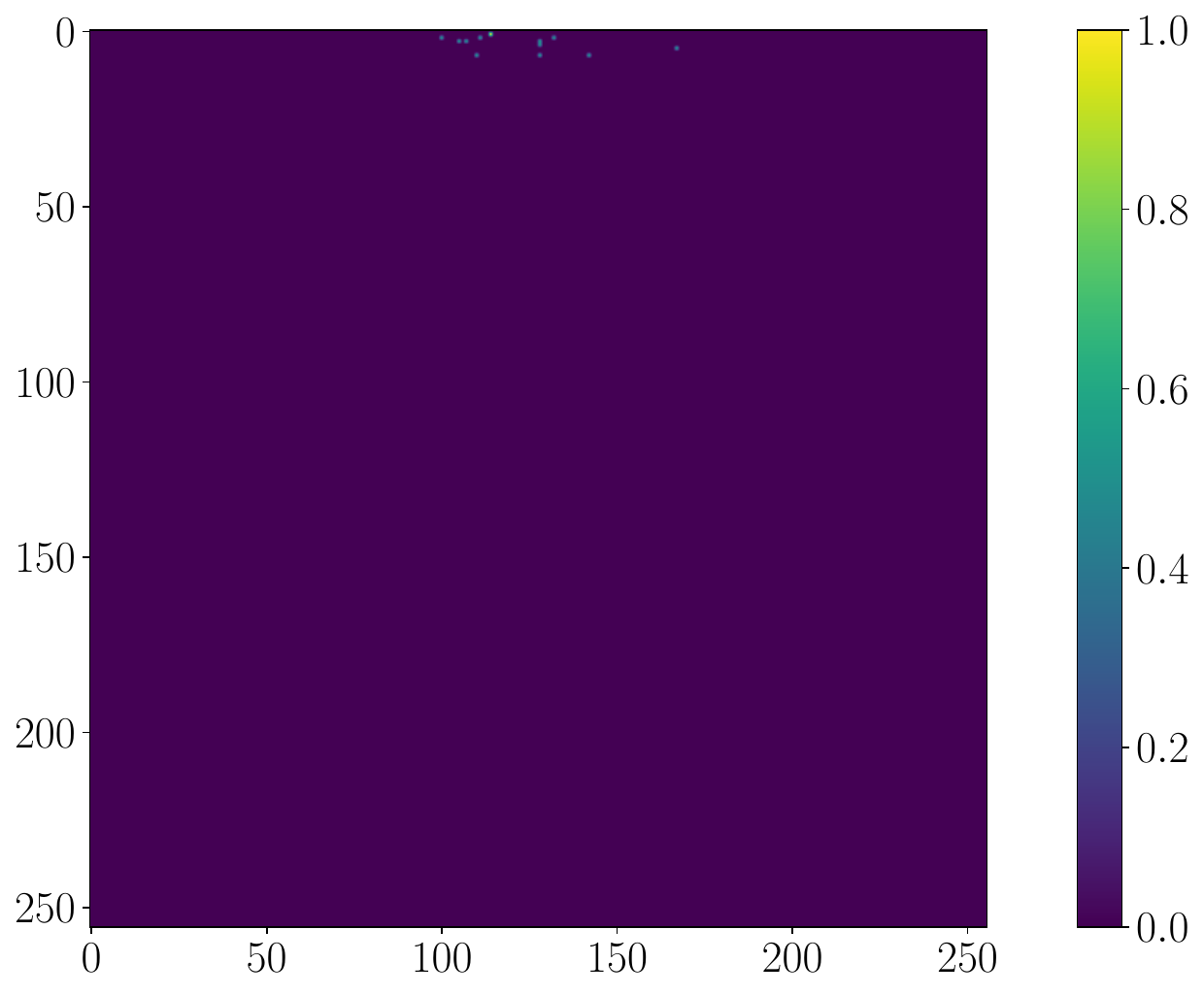}\label{fig:I_background}}
     \subfloat[DoS]{\includegraphics[width=0.27\linewidth]{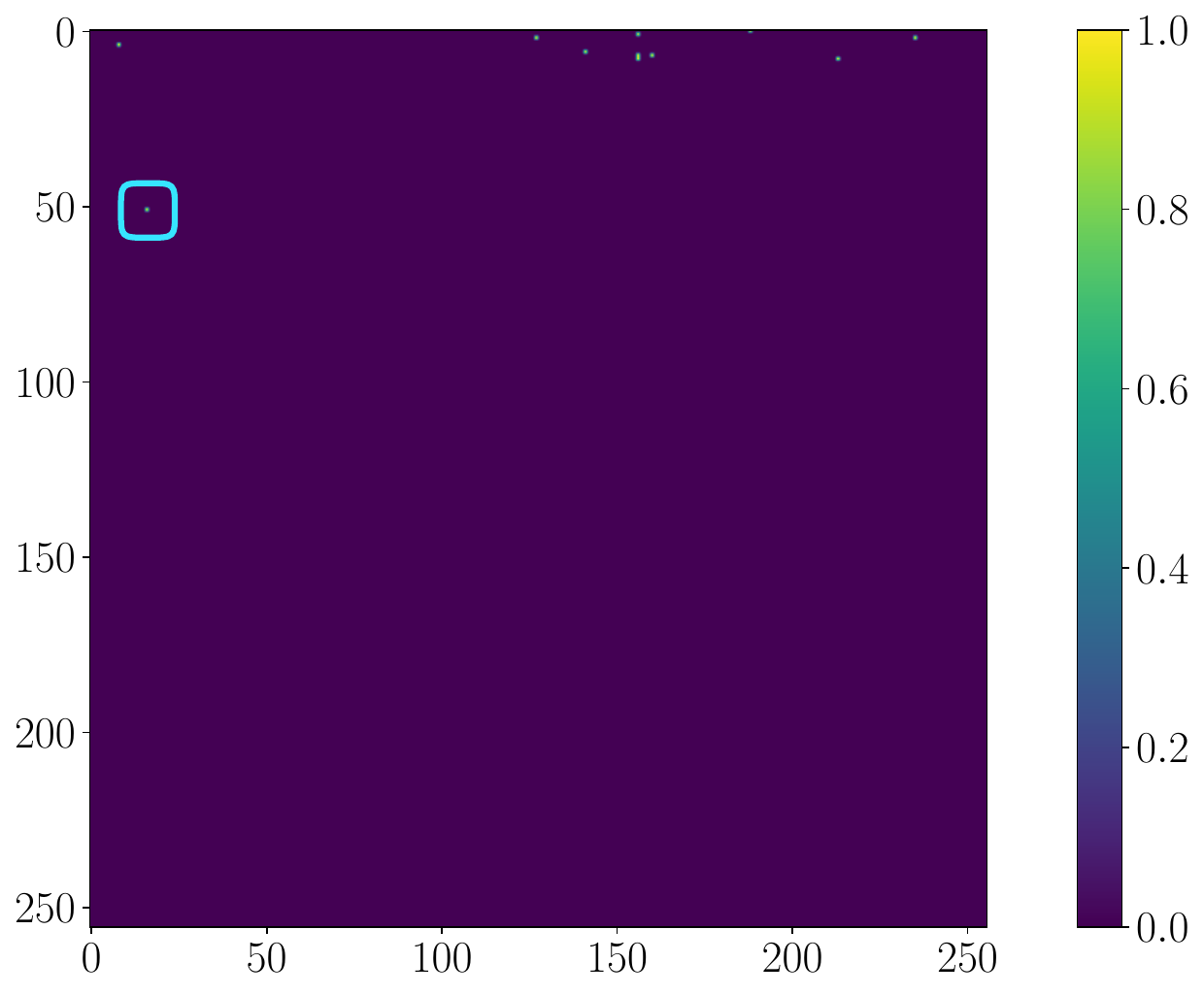}\label{fig:I_dos}}
     \subfloat[Scan]{\includegraphics[width=0.27\linewidth]{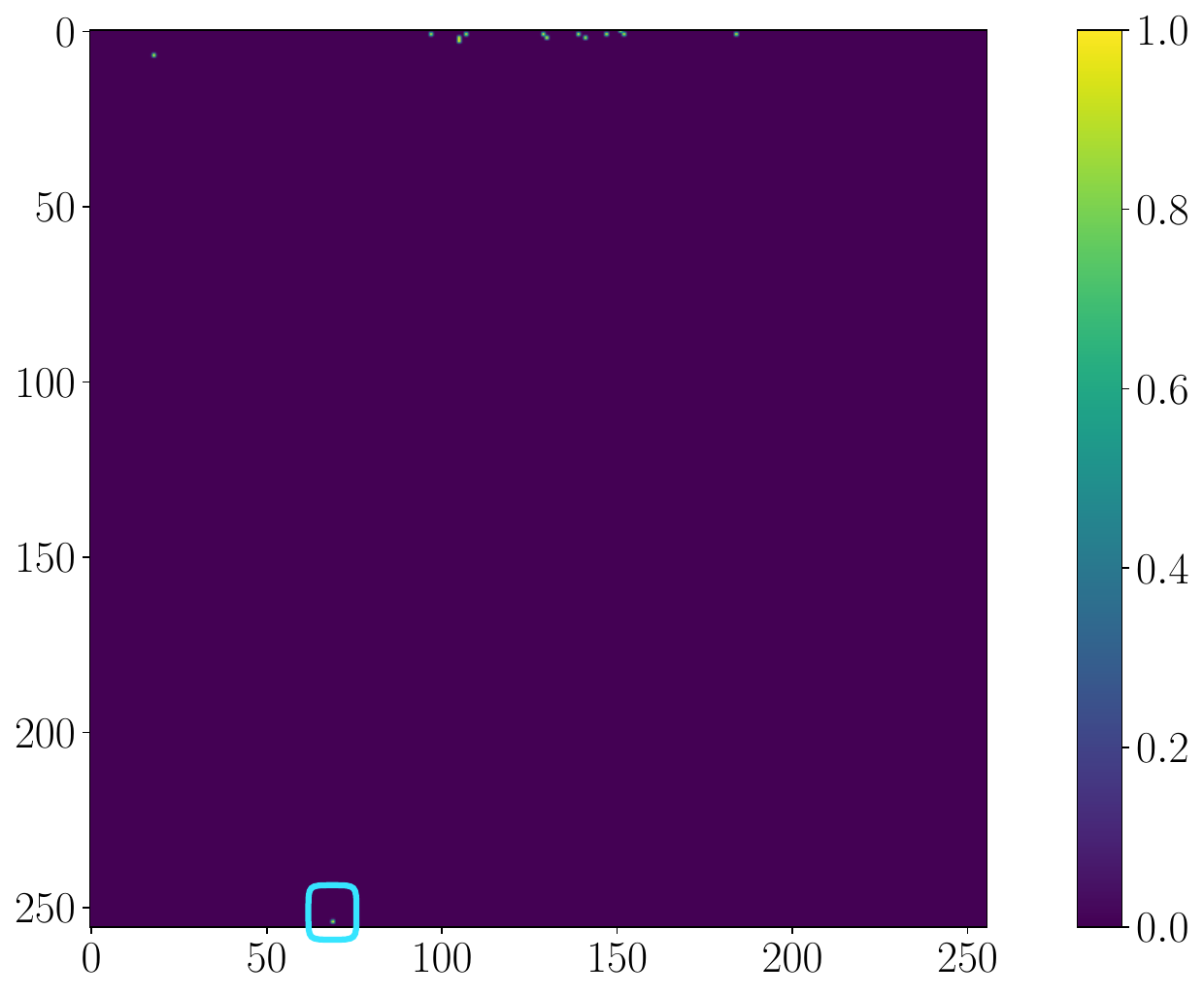}\label{fig:I_scan}}
     \caption{Examples of $\mathcal{I}$ for normal and anomalous traffic. The light-blue square indicates the pixels corresponding to anomalous traffic patterns. The matrices are obtained for 1-second time windows from the dataset presented in~\cite{Macia_2018_Computers}.}
     \label{fig:I}
\end{figure*}
Recently, deep learning in anomaly detection has represented an important shift from traditional \ac{PCA}-based methods. %Unlike \ac{PCA}, which relies on linear transformations, 
Deep learning approaches can capture non-linear relationships and high-level abstractions, offering enhanced detection capabilities in diverse and complex scenarios~\cite{Bovenzi_2023_CompNet}.
%An example of vanilla \ac{AE} for network anomaly detection is in~\cite{Bovenzi_2023_CompNet} where the authors employed packet-level processing for feature extraction and distance metrics for computing the anomality score. 
%Since network traffic is highly correlated in time, recently the research community has started developing recurrent neural networks to exploit the time consistency of network data. In~\cite{Smolen_2023_FRUCT} the authors compared the performance of a \ac{LSTM}-based \ac{AE} with an Isolation Forrest approach for Botnet detection, highlighting the superiority of the former. Following the same design rationale, in~\cite{Zhou_2022_TIFS} a \ac{LSTM}-based \ac{AE} was devised to process log recordings of the network status to model the normality of data flows. 
%To extract graph-oriented information from network traffic, Wang \textit{et al.}~\cite{Wang_2022_TIFS} proposed \textit{Wrongdoing Monitor}, a \ac{GNN} that combines unsupervised and supervised learning to model and detect anomalies from sequential data. Similarly, in~\cite{Miao_2023_Electronics} the authors combined a \ac{GNN} and a \ac{LSTM} \ac{AE} to detect anomalies of traffic flows by means of the reconstruction error.
However, \ac{AE}-based anomaly detectors trained on normal traffic data are prone to generalization problems~\cite{Bovenzi_2023_CompNet}. In fact, even unseen abnormal patterns can be correctly retrieved by reconstruction-based approaches. To mitigate this problem, One-Class \ac{SVM}, \ac{VAE}, and \ac{GAN} have been proposed in literature~\cite{Chen_2024_CN, Fu_2023_WWW, Zhang_TIFS_2023, Geng_2025_CN}. % In~\cite{Zavrak_Access_2020} the superiority of \ac{VAE} with respect to vanilla \ac{AE} has been demonstrated, and One-Class \ac{SVM} have been employed for detecting abnormal patterns of network flow features. In addition, recent works employed both Isolation Forest and \ac{VAE} for improving the detection of malicious network traffic~\cite{Zhang_TIFS_2023}. 
%Regarding adversarial architectures, Fu \textit{et al.}~\cite{Fu_2023_WWW} proposed GANAD, a \ac{GAN}-based approach that encodes network flow features using an \ac{AE}-like structure with a discriminator. Then, by generating new samples from the latent space by the discriminator, the overall architecture was trained in a game-theoretic fashion. Towards this direction, in~\cite{Lunardi_2023_TNSM} a convolutional Wasserstein \ac{GAN} was devised by exploiting a subset of raw bytes of a few initial packets of network flows.

% Recent works that used Transformers
%Lastly, recent works had exploited the well-known Transformer architecture from natural language processing to model normal activity of a network from system logs~\cite{Han_2023_AppliedSciences, Li_2023_AppliedSciences}.

Instead of realizing a complex learning-based detector, we propose to address the network anomaly detection issue from a different perspective. One of the limitations of the existing methods is that they directly process network flows and traffic features without performing a pre-processing step for highlighting hidden traffic peculiarities. This work, on the contrary, proposes an image-based representation of network traffic. The proper definition of the image representation provides a compact picture of the current condition of the monitored network, reducing the complexity of the learning architecture. This direction has been partially investigated in~\cite{Baldoni_Frontiers_2022}, where a 2D representation of network activity for \acp{CPS} has been devised.
However, the defined 2D representation is sparse thus failing in simplifying the processing pipeline.
Indeed, in~\cite{Casarin_SDS_2022} a complex \ac{VAE} has been implemented for detecting anomalies from the representation proposed in~\cite{Baldoni_Frontiers_2022} using reconstruction-based errors for the detection.
%and not suitable as-is for  deep learning processing. 
In this work, we leverage the representation used in~\cite{Baldoni_Frontiers_2022, Casarin_SDS_2022} reducing the image sparsity and making the network patterns more evident. 
This allows to reduce the complexity of the learning architecture. To demonstrate this assertion, we test two reconstruction-based anomaly detectors: a lightweight \ac{VAE} and a vanilla \ac{AE}. 

% Moreover, we compare the anomaly detection performance with recent state-of-the-art papers.
%Hence, this work addresses this issue, introducing a new arrangement for this image-based representation and 
%Moreover, moving from \ac{PCA}-based approaches to variational and adversarial methods, the complexity of the proposed architectures has steadily increased. 
The contributions of this work can be summarized as follows: i) a new image-based representation of network traffic that highlights the presence of attacks, thus requiring a low-complexity anomaly detector; ii) the quantitative comparison between different types of 2D representations; iii) the comparison between anomaly detectors with different complexities. 

\begin{figure}[htb]
     \centering
     \subfloat[Normal, $\mathcal{I}$]{\includegraphics[width=0.23\linewidth]{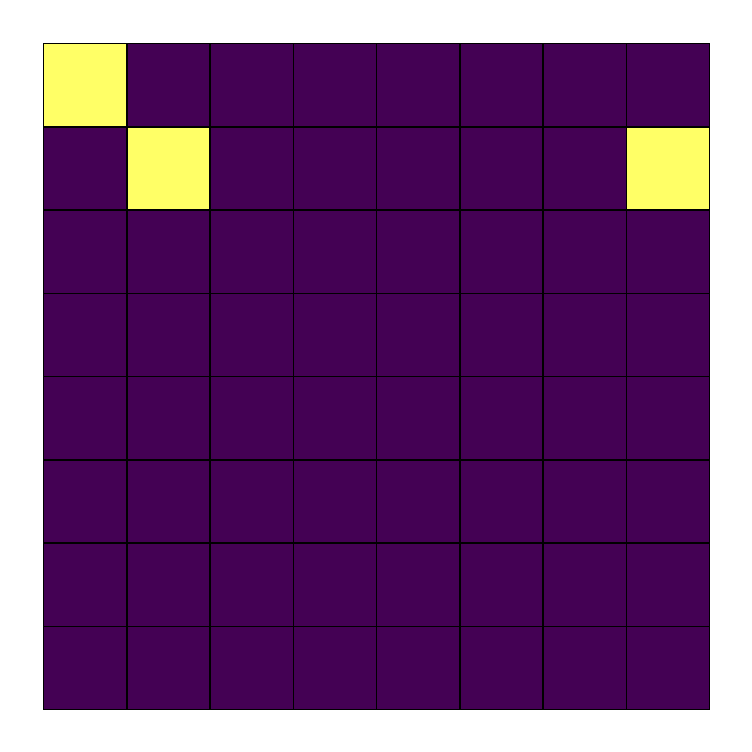}\label{fig:I_background_sample}}
      \hspace{0.1cm}\subfloat[Anomaly, $\mathcal{I}$]{\includegraphics[width=0.23\linewidth]{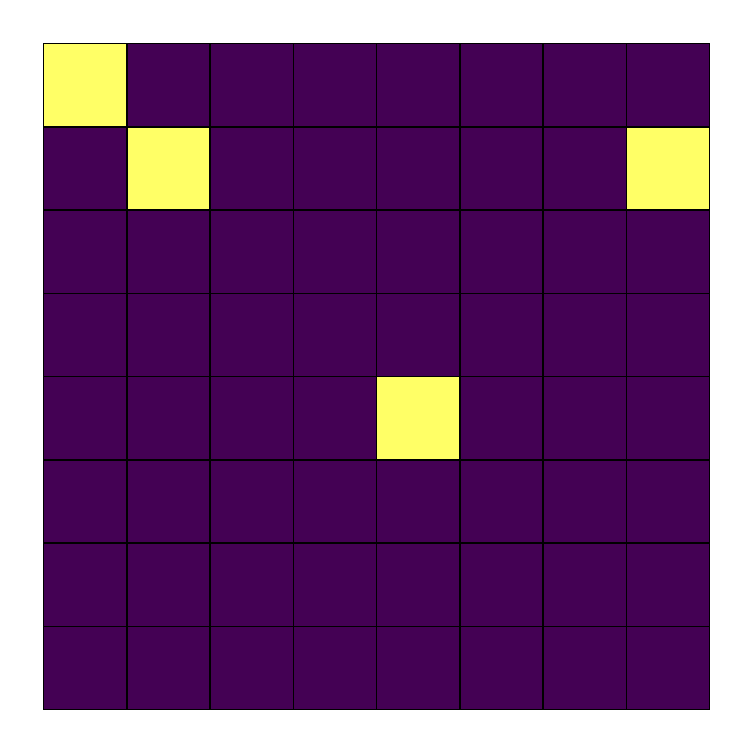}\label{fig:I_anomaly_sample}}
     \hspace{0.1cm}\subfloat[Normal, $\mathcal{I_C}$]{\includegraphics[width=0.23\linewidth]{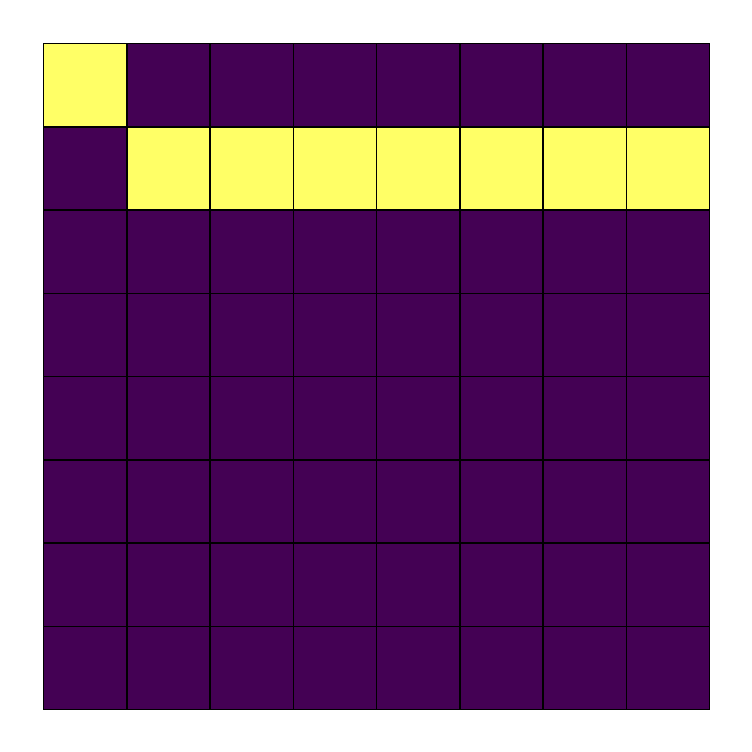}\label{fig:I_C_background_sample}}
     \hspace{0.1cm}\subfloat[Anomaly, $\mathcal{I_C}$]{\includegraphics[width=0.23\linewidth]{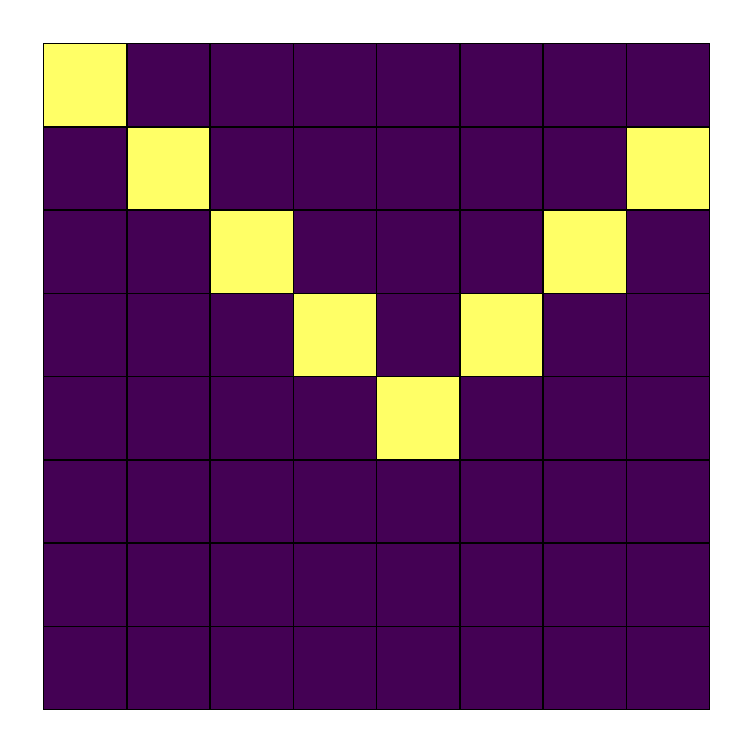}\label{fig:I_C_anomaly_sample}}
     \caption{Examples of $\mathcal{I}$ and $\mathcal{I}_C$ for normal and anomalous traffic in a limit case scenario.}
     \label{fig:sample}
\end{figure}
%The remainder of the paper is organized as follows. Sec.~\ref{sec:Method} introduces the anomaly detection pipeline, describing the proposed network traffic representation, the architectures of the \acp{AE}, and the selection of the threshold. In Sec.~\ref{sec:Results} the dataset and the experimental results are presented and discussed whereas in Sec.~\ref{sec:conclusion} the conclusions are drawn. 
% Works that had used UGR'16 dataset recently 
% \mich{Recent published works that had used UGR for assessing their methods}~\cite{Alghawli_Alexandria_2022, Clausen_Computers_2021, Perez_ToN_2022, Zoppi_TranCPS_2021, Shajari_Access_2022}. \\ 
\begin{figure*}[htb]
     \centering
     \subfloat[Normal]{\includegraphics[width=0.27\linewidth]{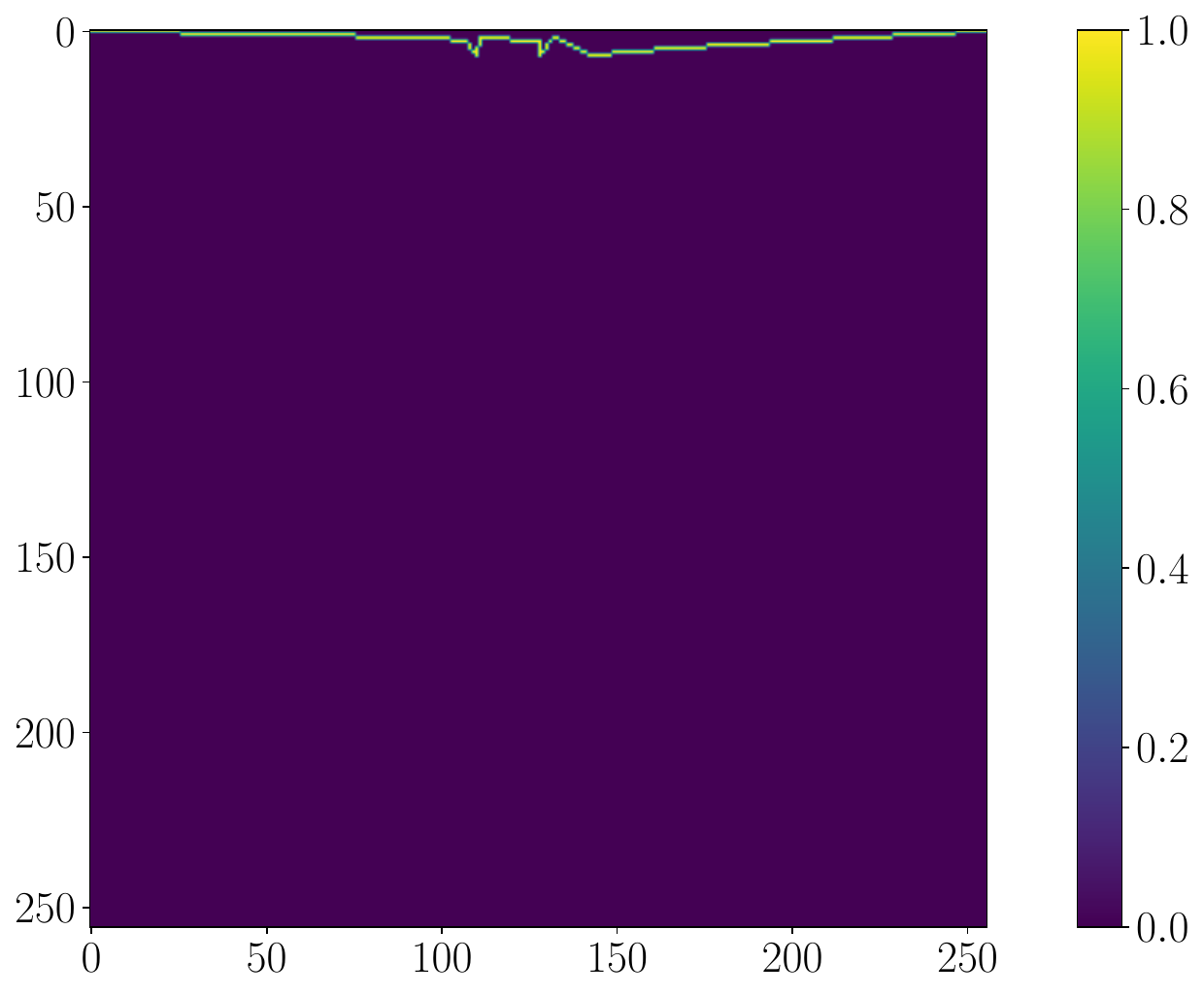}\label{fig:I_C_background}}
     \subfloat[DoS]{\includegraphics[width=0.27\linewidth]{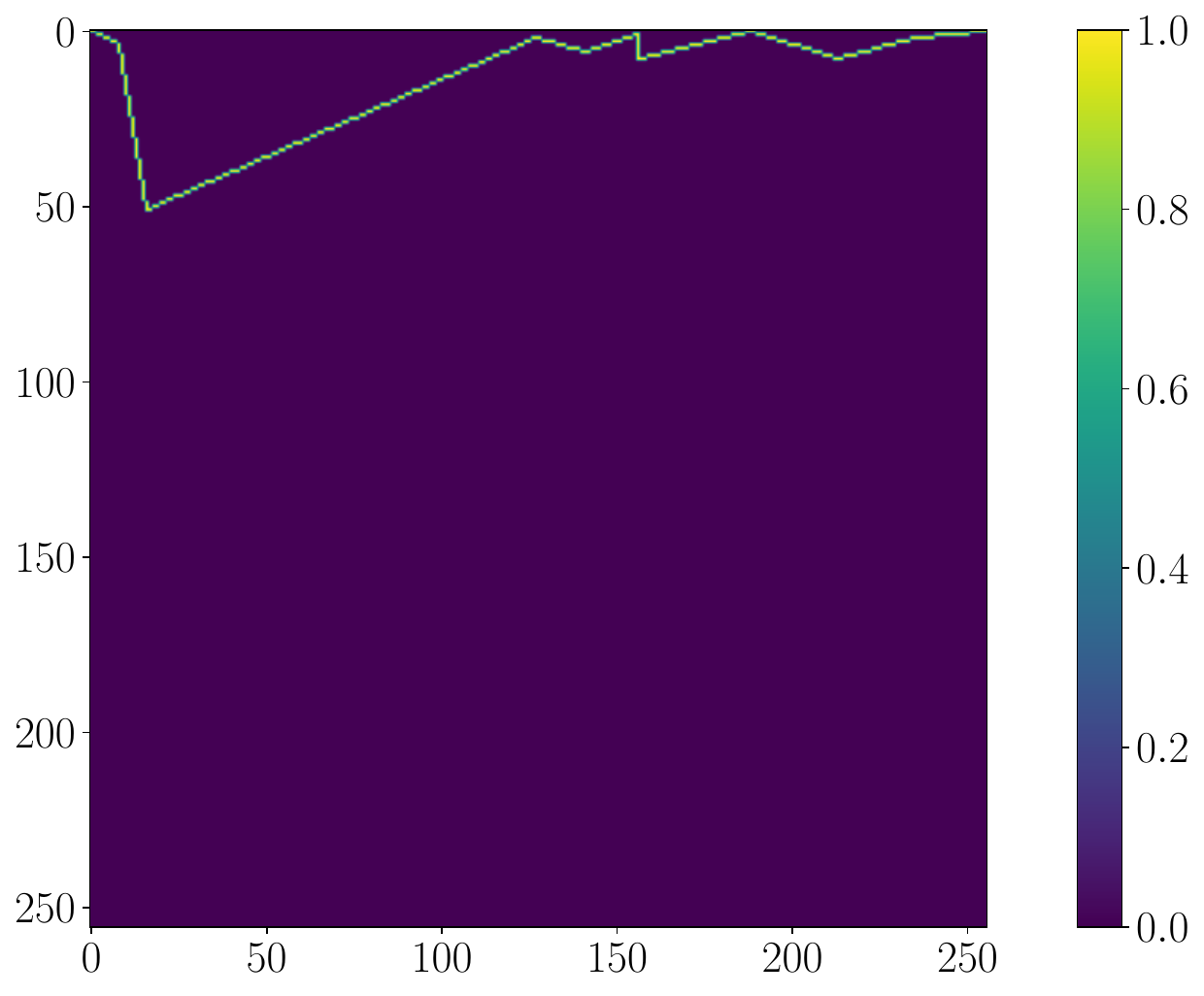}\label{fig:I_C_dos}}
     \subfloat[Scan]{\includegraphics[width=0.27\linewidth]{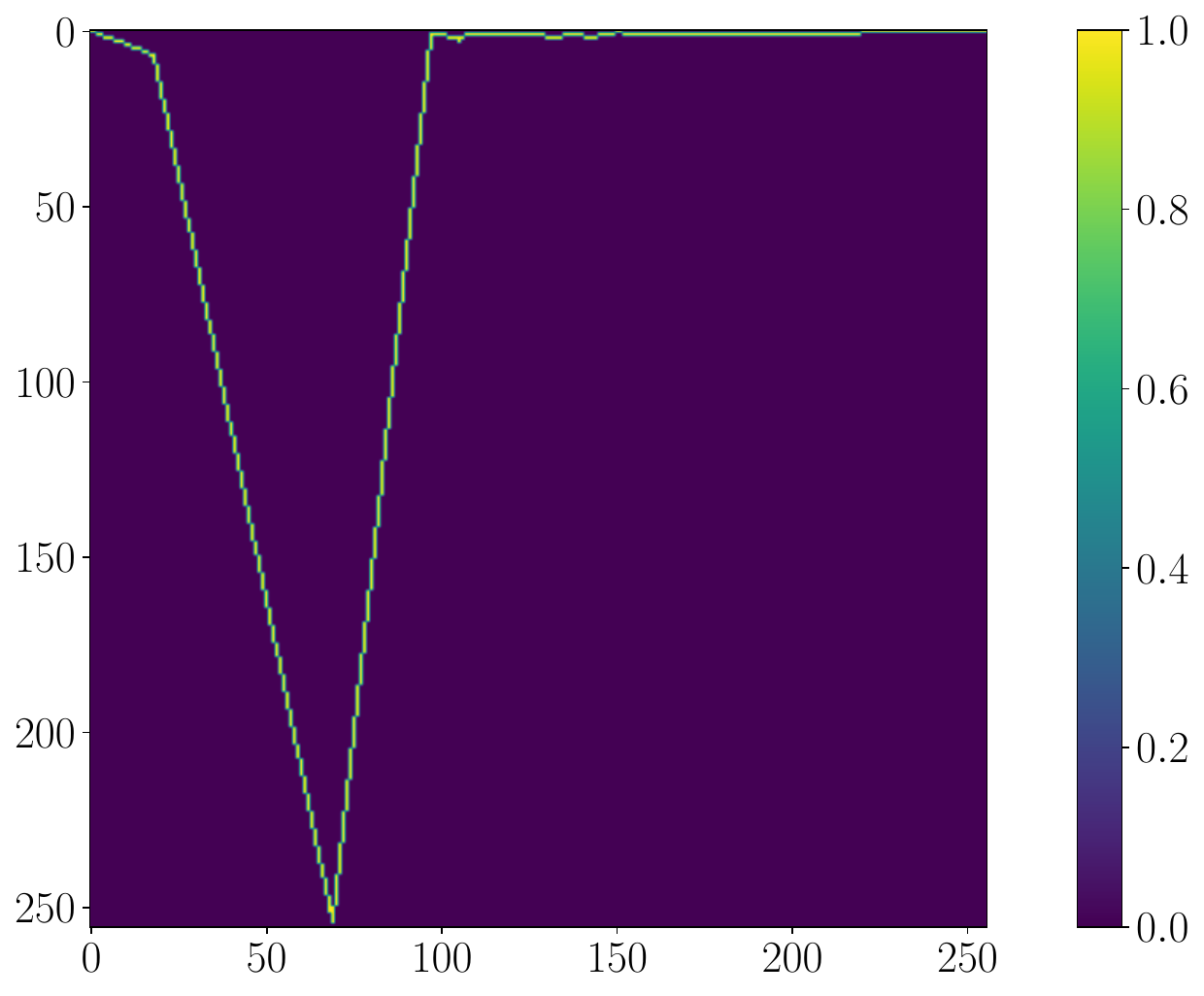}\label{fig:I_C_scan}}
     \caption{Examples of $\mathcal{I}_C$ for normal and anomalous traffic. The matrices are obtained for 1-second time windows from the dataset presented in~\cite{Macia_2018_Computers}.}
     \label{fig:I_C}
\end{figure*}

\section{Anomaly detection method}\label{sec:Method}
In this section an anomaly detector which exploits an image-based traffic representation is described. In more detail, a 2D matrix containing network traffic information has been defined and an \ac{AE} has been designed and trained to reconstruct normal traffic images in an unsupervised fashion. 
By doing so, anomalous traffic representations are poorly retrieved by the \ac{AE}, yielding high reconstruction errors. This property makes the \ac{AE} suitable for anomaly detection scenarios where labeled anomalous instances may be scarce or unavailable. In fact, the model learns to generalize from the normal data without explicitly receiving information about anomalies. This capability represents the reason for which we opted for an \ac{AE}-based strategy.
%After the training process, the selection of the anomaly threshold is performed in order to distinguish between anomalous and normal images.

\subsection{Network traffic representation}\label{sec:NetRappr} 
% Let us consider a network scenario in which $N$ network nodes need to be monitored in order to check for anomalies. The anomaly detection is performed by a control node which collects the data from the monitored sub-network. In order to share the computational load among the different processing units, each node pre-processes the incoming traffic and provides the result to the control node. Thanks to the processing of the information provided by the monitored network, the control node identifies the presence of anomalies. A representation of the considered network scenario is presented in Figure~\ref{fig:network}.

% \begin{figure}[htb]
%  \centering
%  \includegraphics[width=0.9\columnwidth]{Images/network.pdf}
%  \caption{Network representation.}
%  \label{fig:network}
% \end{figure}

% This work aims at defining both the pre-processing procedure performed by each network node, and the anomaly detection method carried out by the control node. More specifically, this work pursues the following goals:
% \begin{itemize}
%     \item the definition of a processing procedure performed by each node aiming at providing to the control node a representation of the incoming traffic which highlights the attack presence;
%     \item the reduction of the communication load between the monitored network and the control node;
%     \item the definition of an anomaly detection algorithm to effectively identify the presence of anomalies. 
% \end{itemize}

The image-based representation proposed in~\cite{Baldoni_Frontiers_2022} associates every pixel to a source-destination IP pair, $(i,j)$, and computes every pixel value, $p(i,j)$, as
\begin{equation}\label{eq:pixel}
    p(i,j)=\frac{\Sigma(i,j)-\mu(i,j)}{\sigma(i,j)+ 10^{-4}}\times e^{f(i,j)},
\end{equation}
where $\Sigma(i,j)$ is the amount of bytes exchanged between $i$ and $j$ in a time window, $\mu(i,j)$ and $\sigma(i,j)$ are the corresponding mean and standard deviation, and $f(i,j)$ is the number of flows.
Although this representation effectively highlights the presence of an attack, it is not suitable for being processed through deep learning algorithms due to the high signal dynamics~\cite{Casarin_SDS_2022}. Therefore, in this work, we adopt the same mapping used in~\cite{Casarin_SDS_2022, Baldoni_CN_2025}. Specifically, every column is associated to a node of the monitored network, and the incoming traffic is processed as in Eq.~\eqref{eq:pixel} and mapped to a different row through a procedure based on the incoming traffic histogram. 
The mapping allows to concentrate the normal traffic in the upper area of the image, while moving the attack-related pixels to lower rows (for the mapping details see~\cite{Baldoni_CN_2025}). 
In the following this representation will be referred to as $\mathcal{I}$.
Fig.~\ref{fig:I} depicts an example of $\mathcal{I}$ in normal conditions and in presence of attacks, considering a time window of $1$ second. As can be noticed, $\mathcal{I}$ is a sparse matrix and
the difference between normal and anomalous images is given by few and isolated points, highlighted in Fig.~\subref*{fig:I_dos} and~\subref*{fig:I_scan}.
%although there is a difference between normal and anomalous images, this difference is represented by few, isolated points. 
This may result in an ineffective autoencoder training due to the high sparsity of active pixels. To highlight this issue, let us consider the limit case in which a single pixel is modified with respect to the image representing normal traffic, as shown in Fig.~\subref*{fig:I_background_sample} and~\subref*{fig:I_anomaly_sample}. 
Under these circumstances, even if the autoencoder provides as output a matrix which is very similar to the normal image (Fig.~\subref*{fig:I_background_sample}), the reconstruction error based on the comparison with the input anomalous matrix (Fig.~\subref*{fig:I_anomaly_sample}) would be very small and the anomaly would not be detected.
To solve this issue, we introduce a new image-based representation, $\mathcal{I}_C$, which can be obtained by connecting all the active pixels in $\mathcal{I}$. To visualize this phenomenon, we reported the $\mathcal{I}_C$ representations corresponding to Fig.~\subref*{fig:I_background_sample} and~\subref*{fig:I_anomaly_sample} in Fig.~\subref*{fig:I_C_background_sample} and~\subref*{fig:I_C_anomaly_sample}. As can be observed, even in the limit case in which the difference between normal and anomalous traffic corresponds to a single pixel modification in $\mathcal{I}$, a relevant difference appears in $\mathcal{I}_C$, thus making the representation less sparse and the training more effective. 
The images $\mathcal{I}_C$ corresponding to the $\mathcal{I}$ matrices represented in Fig.~\ref{fig:I} are reported in Fig.~\ref{fig:I_C}. 
As can be noticed, thanks to the active pixel connection, normal traffic is represented by a signal that oscillates slowly, while the anomalous traffic representation is characterized by spikes. 
%As can be clearly noticed, in this case the differences between normal and anomalous images are significant. 
Finally, it is useful to notice that the computational complexity for obtaining $\mathcal{I_C}$ from $\mathcal{I}$ can be deemed negligible.

\subsection{Autoencoders for anomaly detection}

% To study the performance of learning architectures with different complexities we compare an \ac{AE} and a \ac{VAE}.
The goal of an \ac{AE} is to learn a compressed representation of the input and then reconstruct the original data from it. Anomalies or outliers in the data can be detected by measuring the difference between the original input and its reconstructed version. 
%In this context, background traffic is available, as it it easier to collect rather than abnormal ones~\cite{Claffy_1993_SIGCOMM}, during the training procedure to be correctly retrieved in inference. 
Differently, a \ac{VAE} has also the objective of estimating the true posterior distribution of the latent vectors~\cite{Kingma_2013_ARXIV}. 

% Differently from state-of-the-art approaches~\cite{LopezMartin_Sensors_2017, Xu_WWW_2018, Zavrak_Access_2020, Zhang_TIFS_2023, Casarin_SDS_2022}, a vanilla \ac{AE} is employed rather than a \ac{VAE}. The rationale behind this design choice is to demonstrate the effectiveness of the 2D representation with a simple deep learning architecture. \mich{L'unica motivazione che mi è venuta in mente.}\s{secondo me la motivazione va bene, ma il paragrafo prima lo sposterei nell'intro} \mich{Ora che i risultati sono con il VAE, questo problema non c'è più}

Both architectures are composed of symmetrical encoder $E(\cdot)$ and decoder $D(\cdot)$. The former lossy compresses the input representation with height $H$ and width $W$ $\mathcal{I}_C \in \mathbb{Z}_2^{H \times W}$ to a latent vector $\mathbf{z} \in \mathbb{R}^d$, where $d$ is the dimension of the latent space. If the model is a \ac{VAE}, two latent vectors $\mathbf{z}_{\mu}, \mathbf{z}_{\log \sigma} \in \mathbb{R}^d$ are estimated. In this case, $\mathbf{z} \in \mathbb{R}^d$ is composed by means of the reparametrization procedure~\cite{Kingma_2013_ARXIV}
\begin{equation}\label{eq:1}
\mathbf{z} = \mathbf{z}_{\mu} + \mathbf{z}_{\sigma} + \epsilon, \quad  \epsilon \sim \mathcal{N}(0, 1).
\end{equation}
Consequently, the decoder is responsible for upsampling $\mathbf{z}$ to the reconstructed image $\hat{\mathcal{I}_C}$. The proposed \ac{VAE} and \ac{AE} architectures are detailed in Tab.~\ref{tab:AEStructure}. The encoder $E(\cdot)$ is composed of $3$ convolutional blocks, denoted as $\mathrm{ConvBlock}(c_i)$ where $c_i$ is the number of output channels, that extract spatial features from the input image. Each block performs a 2D convolution, a batch normalization, and the \ac{ELU} activation function. This sequence of operations is employed twice in each block. A $\mathrm{MaxPool}(2,2)$ function is applied after each of the first two convolutional blocks to downsample the images. A linear projection layer is applied to map the output of the last convolutional block into a feature tensor having $1$ channel. This operation is carried out by a $1 \times 1$ convolutional layer. Finally, two fully connected layers, with $n_{fc}$ and $d$ neurons, are responsible for mapping the features into the two latent vectors for means and log variance. 
Symmetrically, $3$ transposed convolutional blocks, indicated as $\mathrm{TranConvBlock}(c_i)$, constitute the decoder $D(\cdot)$. After the first two transposed convolutional blocks, two $\mathrm{UpSample}(2,2)$ operations with nearest neighbour interpolation are performed to retrieve the original shape of the image.
In this work, the number of channels are $\mathbf{c} = [c_1, c_2, c_3] =  [16, 32, 64]$, kernel sizes are $5\times 5$, $n_{fc}$ and $d$ are $128$ and $64$, respectively. This configuration has been tuned by means of hyperparameters optimization algorithms, i.e., grid search. 

Regarding the training procedure, the objective of both architectures is to jointly minimize the error of the encoding-decoding procedure whereas the \ac{VAE} additionally aims to reduce the \ac{ELBO}~\cite{Kingma_2013_ARXIV}. 
The reconstruction error is calculated by comparing the original input and its reconstructed output using a suitable distance or similarity measure. Due to the sparse nature of the image representation, two distance losses are tested. First, the \ac{BCE} is employed as a pixel-wise statistical distance between the input image and its reconstruction. Then, as the images mostly contain few active pixels, a weighted \ac{BCE} loss function is introduced to penalize wrongly reconstructed active pixels with $\mathbf{w} = [w_0, w_1]$, which is the vector that penalizes inactive and active pixels.
\begin{comment}
\begin{equation}\label{eq:wBCE}
    \mathcal{L}_{wBCE}{=} \sum_{i = 1}^{N} \frac{1}{N}  [w_0(\mathcal{I}_{C_i} \log (\hat{\mathcal{I}_{C_i}}) {+} w_1 (1{-} \mathcal{I}_{C_i}) \log (1 {-} \hat{\mathcal{I}_{C_i}}))],
\end{equation}
\end{comment}
%where $N$ is the number of images in a batch, $\hat{\mathcal{I}_{C}} \in \mathbb{R}^{H \times W}$  is the reconstructed image, and  
% If $w_0 = w_1$, then Eq.~\eqref{eq:wBCE} is equal to the unweighted \ac{BCE} $\mathcal{L}_{BCE}$.

Regarding the \ac{VAE}, thanks to the reparameterization in Eq.~\eqref{eq:1}, it is possible to derive the Kullback-Lieber divergence loss as $\mathcal{L}_{KL}(\mathbf{z}, \mathcal{I}_{C}) = D_{KL}( q(\mathbf{z} | \mathcal{I}_{C}) || p(\mathbf{z})),$
where $q(\cdot)$ and $p(\cdot)$ are the learned probability distribution over the latent space and a predefined prior distribution, which is a standard normal distribution, respectively.  
Finally, the total \ac{ELBO} loss employed for training the architecture is $ \mathcal{L}_{ELBO}(\mathbf{z}, \mathcal{I}_{C}, \hat{\mathcal{I}_{C}}) = \mathcal{L}_{wBCE}(\mathcal{I}_{C}, \hat{\mathcal{I}_{C}}) - \beta \mathcal{L}_{KL}(\mathbf{z}, \mathcal{I}_{C})$,
where $\beta = 0.00005$ is set for balancing the two losses' magnitude.

% The \ac{VAE} architecture is the same except for the dense layers that compute $\mathbf{z}_{\mu}$ and $\mathbf{z}_{\log \sigma}$.
\begin{table}[ht]\caption{Description of both \ac{AE} and \ac{VAE} structures.} 
\centering
\renewcommand*{\arraystretch}{1.2}
\adjustbox{max width=0.48\textwidth}{%
\begin{tabular}{c|c}
\textbf{Input}: input image $\mathcal{I}_{C} \in \mathbb{Z}_2^{W \times W}$ & \textbf{Input}: latent vector $\mathbf{z} \in \mathbb{R}^d$  \\
 \hline \hline
 \textbf{Encoder} $E(\cdot)$ & \textbf{Decoder} $D(\cdot)$ \\
 \hline
 $\mathrm{ConvBlock}(c_1)$ & $\mathrm{FC}(n_{fc})$ \\
 $\mathrm{MaxPool}(2,2)$ & $\mathrm{ELU}$ \\
 \cline{1-1}
 $\mathrm{ConvBlock}(c_2)$ & $\mathrm{FC}(n_{fc})$ \\
 $\mathrm{MaxPool}(2,2)$ & Unflatten to 3D tensor \\
 \cline{1-1}
 $\mathrm{ConvBlock}(c_3)$ & Projection to $C_3$ channels \\
  \cline{2-2}
 Projection to $1$ channel & $\mathrm{TranConvBlock}(c_3)$ \\
 \cline{1-1}
 Flatten to 1D vector & $\mathrm{UpSample}(2,2)$ \\
  \cline{2-2}
 $\mathrm{FC}(n_{fc})$ & $\mathrm{TranConvBlock}(c_2)$ \\
 $\mathrm{ELU}$ & $\mathrm{UpSample}(2,2)$ \\
 \cline{2-2}
 $\mathrm{FC}(d)$ & $\mathrm{TranConvBlock}(c_1)$\\
 \textbf{if} \ac{VAE} \textbf{then} Reparameterization (Eq.~\eqref{eq:1}) & Projection to $1$ channel \\ 
\hline \hline
\textbf{Output}:
latent vector $\mathbf{z} \in \mathbb{R}^d$ & \textbf{Output}: reconstructed image $\hat{\mathcal{I}_{C}}\in \mathbb{R}^{H \times W}$  \\

\end{tabular}}
\label{tab:AEStructure} 
\end{table}

\begin{comment}\subsection{Threshold selection}
To compute the anomaly detection threshold, the trained model has been applied to a validation set composed of unseen normal traffic data. By doing so, it is possible to evaluate how the architecture has modeled non-anomalous traffic. Ideally, the anomaly detector should provide separated reconstruction error distributions for anomalous and normal traffic. However, due to the intrinsic unpredictability of anomalous traffic patterns, no assumptions can be made on the anomaly reconstruction error distribution. Hence, a probability of false alarm $p_{fa}$ is introduced by exploiting the distribution of background errors in the validation dataset. In more details, the anomaly threshold $\delta$ is obtained by selecting the error reconstruction that satisfies the following disequality: 
\begin{equation}
    \mathbb{P}[\mathcal{L}(\mathcal{I}_{C}, \hat{\mathcal{I}_{C}}) < \delta | \mathcal{I}_{C} \in \mathcal{V}] \leq p_{fa}
\end{equation}
where $\mathbb{P}[\cdot]$ evaluates the probability of an event, $\mathcal{L}$ is a generic distance measure, and $\mathcal{V}$ is the set of validation normal samples. As an example, if $p_{fa} = 0$, then $\delta$ is equal to the maximum background error obtained from validation samples. This process allows to select a stricter or looser threshold by decreasing or increasing $p_{fa}$, respectively. 
\end{comment}

% In this work, thanks to this procedure, $p_{fa}$ has been set to $0.5\%$ to balance the trade-off between number of false positives and false negatives.

\section{Results}\label{sec:Results}
The proposed image representation and anomaly detector are tested on the UGR'16 dataset~\cite{Macia_2018_Computers}. The performance of the learning architectures trained on the proposed 2D representation is assessed on the test set by means of \ac{IR} metrics such as precision, recall, accuracy, and F1 score. Moreover, a quantitative analysis is performed, varying the deep learning architecture, the 2D representation, and the loss function during the training phase. 

\subsection{Dataset}
%To assess the performance of the proposed approach, a large-scale dataset that encompass both background and anomalous traffic data is required. In this work, UGR'16 dataset~\cite{Macia_2018_Computers} has been selected and the reasons of this choice are manifold. 
The UGR'16 dataset consists of 
%traffic generated by a wide array of applications, including web browsing, email communication, file transfer, streaming, and peer-to-peer networks. By mirroring actual network usage, the dataset provides a realistic foundation for training anomaly detection models. In more detail, UGR'16 dataset includes 
two subsets: calibration and test. The calibration capture consists of real background network traffic and can be employed for training normality models. Instead, the test capture includes both clean traffic and anomalous flows obtained as the combination of background traffic and controlled attack traffic generated using advanced hacking tools. Three classes of attack have been considered: \ac{DoS}, scan, and botnet. Since the effect of botnet over the normal traffic has not been taken into account in~\cite{Macia_2018_Computers}, we selected only the first two attack categories.
These attacks are denoted as DoS53, DoS11, Scan44, and Scan11. The first number in the attack names refers to the number of attackers whereas the second identifies the number of victims~\cite{Macia_2018_Computers}. In this work, $320,000$ images have been generated from the calibration set with a $1$-second time window. The validation set is obtained by sampling $10\%$ of the training set. For testing, $70,462$ images have been generated from the test set. $60,000$ images are normal, $1,676$ represent DoS11, $4,596$ are DoS53 attack samples, $992$ are Scan11 attacks, and $3,198$ images are Scan44 representations. Based on the network structure in the dataset, images $\mathcal{I}_{C}$ are of shape $H = 256$ and $W = 256$. 

% Further details about the attack procedures are available in~\cite{Macia_2018_Computers}.

%To test anomaly detection algorithms that consider cyclostationarity, UGR'16 dataset has been select because a standardized batch of attacks is defined and executed at different times and days to examine if detector accuracy varies based on time or day.

\subsection{Results on all attacks}

Tab.~\ref{tab:allAttacks} depicts the performance of the proposed approach on all the considered attacks with different options for the loss function and the learning architecture. It is worth noticing the superiority in performance of $\mathcal{I}_C$ with respect to $\mathcal{I}$. 
More specifically, the \ac{VAE}, both with unweighted and weighted \ac{BCE}, suffered from mode collapse using $\mathcal{I}$, i.e., all the input images have been mapped to the same latent vector, providing poor generalization capabilities. 
Concerning the \ac{AE}, to achieve a recall of approximately $50\%$ when using $\mathcal{I}$, the $p_{fa}$. i.e., probability of false alarm, threshold has to be set to $20\%$, thus proving that $\mathcal{I}$ is not a suitable input for low-complexity deep learning algorithms. In addition, the results obtained using $\mathcal{I}_C$ with unweighted \ac{BCE} clearly show the importance of the introduced weighting procedure to penalize wrongly reconstructed active pixels. 
To fairly compare the performances of the \ac{AE} and the \ac{VAE} when using $\mathcal{I}_C$, we selected as target $p_{fa}$ a value of $0.15\%$. Tab.~\ref{tab:allAttacks} clearly indicates that, thanks to the definition of $\mathcal{I}_C$, the \ac{AE} and the \ac{VAE} achieve comparable performances, although the former has a higher precision and the latter has a higher recall.
\begin{table}[ht]\caption{Results on all attacks. Dash symbol means random predictions, i.e., the model suffered from mode collapse.}
\centering
    \adjustbox{max width=0.45\textwidth}{
    \renewcommand*{\arraystretch}{1.2}
\begin{tabular}{cccc|cccc}
\hline \hline
Model & wBCE & Image & $p_{fa}$ (\%) & F1 & Recall & Precision & Accuracy \\
\hline \hline
\ac{AE}  & $1$ &  $\mathcal{I}$ & $20$ & $0.3662$ & $0.4551$ & $0.3072$ & $0.7678$ \\
\ac{AE}  & $15$ & $\mathcal{I}$ & $20$ & $0.3897$ & $0.4730$ & $0.3302$ & $0.7795$ \\
\ac{AE}  & $1$ &  $\mathcal{I}_C$ & $14$ & $0.8058$  & $0.8835$ & $0.7407$  & $0.9367$ \\
\rowcolor{Gray} \ac{AE}  & $15$ & $\mathcal{I}_C$ & $0.15$ & $0.9405$  & $0.9093$  & $0.9739$  & $0.9829$  \\
\hline
\ac{VAE} & $1$ &  $\mathcal{I}$ & - &  - & -  &  - & - \\
\ac{VAE} & $15$ & $\mathcal{I}$ & - &  - & -  &  - & - \\
\ac{VAE} & $1$ &  $\mathcal{I}_C$ & $0.15$ & $\mathbf{0.9428}$ & $0.9037$ & $\mathbf{0.9852}$ & $\mathbf{0.9837}$ \\
\rowcolor{Gray} \ac{VAE} & $15$ & $\mathcal{I}_C$ & $0.15$ & $0.9248$ & $\mathbf{0.9233}$ & $0.9264$ & $0.9772$  \\     
\hline \hline
\end{tabular}
}
\label{tab:allAttacks}
\end{table}

To better understand which attacks are easier to detect exploiting the two approaches, Tab.~\ref{tab:attackwise} shows the performance of the \ac{AE} and \ac{VAE} with $p_{fa}=0.15\%$ on each attack scenario. These results have been obtained considering as test set the combination of background test data and the single attack samples. 
Overall, the detection performance on \ac{DoS} are better with respect to scan attacks. A possible interpretation is that, on average, the $\mathcal{I_C}$ representation is less effective in highlighting the presence of scan attacks, thus resulting in images that resemble normal traffic which are not easily detected by lightweight architectures.
It is worth highlighting how the \ac{AE} is more precise to detect DoS53 than the \ac{VAE} while the latter outperforms the former in identifying Scan11 attacks in terms of recall, albeit with a higher rate of false positives. 
%To visualize the behavior of the anomaly detectors, Figure~\ref{fig:tsne_hist} shows the t-SNE~\cite{VanDerMaaten_2008_JMLR} and the histogram of anomalous and clean traffic samples using the \ac{VAE} and the \ac{AE} trained on $\mathcal{I}_C$. It is worth noting that both t-SNE plots contain distinct clusters encompassing the two types of attacks (\ac{DoS} and scan) and the normal samples. 
To further compare the two architectures, we computed the number of parameters to learn as well as the number of \acp{GMAC}.
%As stated in~\cite{Zhang_2021_CommunicationsACM}, learning-based models with fewer parameters require less computational power and time during both the optimization and inference phases, reducing response time of the detection system. 
Both models have $1.2$ millions of parameters, and $1.84$ \acp{GMAC}, thus being equally lightweight~\cite{Zhang_2021_CommunicationsACM}.
Therefore, it is possible to conclude that the proposed pre-processing of the incoming traffic allows to effectively and promptly detect the presence of attacks with low-complexity learning architectures. Depending on the specific application for which the anomaly detector is employed, the learning model can be selected based on the performance metric that needs to be prioritized.

% In fact, although the \ac{VAE} achieves better performances, it provides only a $1.9$\% of Recall improvement with respect to its non-variational counterpart. 

% This behaviour demonstrates the ability of the models not only to identify attacks in an unsupervised manner but also to distinguish which class of attack is present in a network within $1$ second. 

% \s{Io aggiungerei un commento sulle pfa, inoltre a parità di pfa l'AE è del tutto comparabile con il VAE - io rigirerei tutto il discorso generale del paper non come la proposta di un VAE ma come il confronto tra AE e VAE mostrando che quando l'input è sensato non serve tutta quella complessità computazionale e questo prova in modo schiacciante la superiorità della rappresentazione rispetto a SDS} 

\begin{comment}
\begin{figure}[htb]
 \centering
 \includegraphics[width=1\columnwidth]{Images/AllAttacks.pdf}
 \caption{Experimental results on UGR'16 on all attacks.}
 \label{fig:allAttacks}
\end{figure}
\end{comment}
% It is worth highlighting the difficulty of all the detectors for the DOS53 attack (\mich{come mai?}) and the simplicity of detecting DOS11 attacks \s{Io questa cosa non la metterei, perché in realtà non c'è nessun buon motivo per cui dovrebbe accadere}
\begin{table}[ht]\caption{Attack-wise results of the two anomaly detectors in gray from Tab.~\ref{tab:allAttacks}. All methods employ $\mathcal{I}_C$ as input.}
\centering
    \adjustbox{max width=0.45\textwidth}{%
\begin{tabular}{c|ccccccc}
\hline \hline
Attack & Model & wBCE & F1 & Recall & Precision & Accuracy \\ \hline
\multirow{2}{*}{DoS11}  & AE       & $15$   & $\mathbf{0.9243}$ & $0.9899$ & $\mathbf{0.8668}$ & $\mathbf{0.9956}$          \\ 
                        & VAE      & $15$   & $0.8033$ & $\mathbf{0.9905}$ & $0.6756$ & $0.9870$          \\ \hline \hline
\multirow{2}{*}{DoS53}  & AE       & $15$   & $\mathbf{0.9697}$ & $\mathbf{0.9934}$ & $\mathbf{0.9471}$ & $\mathbf{0.9956}$         \\  
                        & VAE      & $15$   & $0.9165$ & $0.9928$ & $0.8510$ & $0.9872$          \\ \hline \hline
\multirow{2}{*}{Scan11} & AE       & $15$   & $\mathbf{0.7647}$ & $0.7782$ & $\mathbf{0.7517}$ & $\mathbf{0.9922}$         \\   
                        & VAE      & $15$   & $0.6368$ & $\mathbf{0.8387}$ & $0.5132$ & $0.9844$          \\ \hline \hline
\multirow{2}{*}{Scan44} & AE       & $15$   & $\mathbf{0.8432}$ & $0.7871$ & $\mathbf{0.9080}$ & $\mathbf{0.9852}$         \\   
                        & VAE      & $15$   & $0.7893$ & $\mathbf{0.8183}$ & $0.7623$ & $0.9779$        \\ \hline \hline
\end{tabular}
}
\label{tab:attackwise}
\end{table}

\begin{table}[htb]\caption{Comparison in terms of recall with state-of-the-art on UGR'16. Dash symbol $-$ means not available.}
\centering
\adjustbox{max width=1\linewidth}{%
\begin{tabular}{cccccccc}
\hline \hline
\multicolumn{2}{c}{Approach}                                                      & DoS11  & DoS53  & DoS             & Scan11          & Scan44          & Scan            \\ \hline
\multicolumn{1}{c|}{\multirow{4}{*}{S}}   & LR~\cite{Carrion_AS_2020}                    & -      & -      & $0.9150$          & -               & -               & {\ul $0.9160$}    \\
\multicolumn{1}{c|}{}                              & RF~\cite{Carrion_AS_2020}                    & -      & -      & $0.8840$          & -               & -               & $\mathbf{0.9250}$ \\
\multicolumn{1}{c|}{}                              & Radial \ac{SVM}~\cite{Carrion_AS_2020}               & -      & -      & $0.8310$           & -               & -               & $0.5360$          \\
\multicolumn{1}{c|}{}                              & Linear \ac{SVM}~\cite{Carrion_AS_2020}                 & -      & -      & $0.8980$          & -               & -               & $0.9100$          \\ \hline
\multicolumn{1}{c|}{\multirow{2}{*}{U}} & Kitsune~\cite{Medina_FQAS_2023}               & -      & -      & $0.6400$          & $0.0100$          & $0.7200$          & -               \\
\multicolumn{1}{c|}{}                              & Tensor-based~\cite{Shajari_Access_2022} & -      & -      & $\mathbf{0.9966}$ & $\mathbf{0.9999}$ & $\mathbf{0.9999}$ & -               \\
\rowcolor{Gray}\multicolumn{1}{c|}{}         & AE                           & $0.9900$ & $0.9935$ & {\ul $0.9927$}    & $0.7782$          & $0.7871$          & $0.7849$          \\
\rowcolor{Gray}\multicolumn{1}{c|}{\multirow{-2}{*}{Ours}}                              & VAE                          & $0.9905$ & $0.9928$ & $0.9925$          & {\ul $0.8387$}    & {\ul $0.8183$}    & $0.8230$          \\ \hline\hline
\end{tabular}}\label{tab:comparison}
\end{table}
\begin{comment}
\begin{figure*}[htp]
\centering
\includegraphics[width=.49\textwidth]{Images/DOS11.pdf}\quad
\includegraphics[width=.49\textwidth]{Images/DOS53.pdf}\quad
\medskip
\includegraphics[width=.49\textwidth]{Images/SCAN11.pdf}\quad
\includegraphics[width=.49\textwidth]{Images/SCAN44.pdf}\quad
\caption{Experimental results on UGR'16 for each attack.}
\label{fig:results}
\end{figure*}
\mich{DOS53 è il più difficile da identificare (ed eventualmente capire il perché). Da mostrare qualche esempio.}
\end{comment}
% Works that had used UGR'16 dataset recently 
% \mich{Recent published works that had used UGR for assessing their methods} \cite{Alghawli_Alexandria_2022, Clausen_Computers_2021, Perez_ToN_2022, Zoppi_TranCPS_2021, Shajari_Access_2022, Camacho_TIFS_2019}

\begin{comment}
\begin{figure*}[htb]
     \centering
     \subfloat[AE]{\includegraphics[width=0.49\linewidth]{Images/tsne_and_histc_AE.png}}\label{fig:tsne_hist_ae}
     \subfloat[VAE]{\includegraphics[width=0.49\linewidth]{Images/tsne_and_histc_VAE.png}\label{fig:tsne_hist_vae}}
     \caption{t-SNE of latent vector $\mathbf{z}$ and histogram of errors of normal and anomalous samples from \ac{AE} and \ac{VAE}, respectively.}
     \label{fig:tsne_hist}
\end{figure*}
\end{comment}

% \begin{figure*}[ht]
% \centering
% \includegraphics[width=1\linewidth]{Images/tsne_and_histc.pdf}
% \caption{t-SNE of latent vector $\mathbf{z}$ and histogram of errors of normal and anomalous samples.}
% \label{fig:tsne_hist}
% \end{figure*} 

\subsection{Comparison with state-of-the-art}
% A direct comparison with respect to state-of-the-art approaches is challenging due to the different windowing of traffic data. Moreover, a generalised metric for network anomaly detection has not been defined yet in literature. However, it is important to highlight that the proposed method:
% \begin{itemize}
%     \item computational complexity
%     \item response time
% \end{itemize}

% \begin{table}[ht]\caption{Performance comparison with state-of-the-art. Dash symbol $-$ indicates that the percentage is not available.}
% \centering
%     \adjustbox{max width=0.48\textwidth}{%
% \begin{tabular}{cccc|c}
% \hline \hline
% Approach & Representations & \# of Params (M) & $p_{fa}$ & Recall \\
% \hline \hline
% \rowcolor{Gray} \ac{AE} & $\mathcal{I}_C$ & $\mathbf{1.2}$M & $0.15\%$ & $0.995$ \\
% \rowcolor{Gray} \ac{VAE} & $\mathcal{I}_C$ & $\mathbf{1.2}$M & $0.15\%$ & $0.990$ \\  
% \hline
% \ac{VAE} from \cite{Casarin_SDS_2022} & $\mathcal{I}_{SoA}$ & $268$M & $-$ & $1.00$ \\
% \hline \hline
% \end{tabular}
% }
% \label{tab:comparison}
% \end{table}

To prove the effectiveness of the proposed representation, we compared our approach with both supervised (S) and unsupervised (U) methods presented in the literature. We report the results in terms of recall in Tab.~\ref{tab:comparison}. 
The proposed method outperforms supervised approaches, i.e., \ac{LR}, \ac{RF}, and \ac{SVM}, for \ac{DoS}, while it achieves lower performance on scan. Differently from supervised methods, the proposed detector has no previous knowledge about the attack types, thus making the anomaly detection more challenging. % Moreover, with respect to supervised techniques, we expect the proposed approach to be more robust against new and unforeseen attack types.
As for the unsupervised methods, our representation achieved significantly higher performance on both attacks with respect to the method presented in~\cite{Medina_FQAS_2023}. In contrast, it reached lower recall values than the technique introduced in~\cite{Shajari_Access_2022}. The performance gap is negligible for DoS, while it is considerable for scan. This behavior can be attributed to the difference in complexity of the two approaches. Although in~\cite{Shajari_Access_2022} the number of parameters and \acp{GMAC} are not provided, they perform 3D convolutions with $64\times64\times64$ input tensors, while we employ 2D convolutions with $256\times256$ images. Moreover, while in~\cite{Shajari_Access_2022} data was processed in $1$-minute chunks, the proposed approach has a significantly higher promptness by using $1$-second time windows. % Therefore, our goal of reducing the complexity of the anomaly detector thanks to the usage of a powerful representation can be considered achieved.
Additionally, we compare our results with the method proposed in~\cite{Casarin_SDS_2022}. To perform a fair comparison, however, the same  data cleaning procedure introduced in~\cite{Casarin_SDS_2022} has been applied. Specifically, the authors retained only the traffic windows for which the mapping was successful (i.e., the  attack pixel $p$ was mapped to a raw index larger than $35$), resulting in $9,541$ anomalous data samples. 
%to separate the performances of the anomaly detector from the ones of the mapping. More specifically, only the traffic windows for which the attack pixel $p$ was mapped to a raw index larger than $35$ were retained, 
%Therefore, we evaluated the performance of the the proposed representation after the same cleaning procedure. 
In~\cite{Casarin_SDS_2022}, a binary classification between normal and anomalous instances has been performed achieving a recall of approximately $1$. By using the proposed representation $\mathcal{I_C}$, we obtained a recall of $0.9951$ and $0.9947$ for the \ac{AE} and \ac{VAE}, respectively. This demonstrates the effectiveness of the proposed representation which, thanks to its ability of highlighting the presence of attacks, leads to similar performance with respect to~\cite{Casarin_SDS_2022} despite the smaller complexity. In fact, while the proposed vanilla architectures require the training of $1.2$M parameters, the \ac{VAE} proposed in~\cite{Casarin_SDS_2022} involves $268$M parameters. Our architectures, instead, are noticeably more lightweight than the one proposed in~\cite{Casarin_SDS_2022} ($1.84$ \acp{GMAC} versus $2.41$ \acp{GMAC}). Therefore, the usage of the $\mathcal{I}_C$ representation allows to decrease the computational complexity of both training and inference processes while achieving similar performance.

\section{Conclusions}\label{sec:conclusion}
In this work, an image-based representation of network traffic has been presented, and its efficacy for anomaly detection has been demonstrated. 
%The proposed approach allows the prompt detection of network anomalies while providing a compact representation of the network conditions. 
The comparison between two lightweight learning architectures and state-of-the art approaches highlighted that, thanks to the definition of a proper image-based representation, it is possible to reduce the computational complexity of the processing algorithm. A possible drawback of the proposed approach is that 1-second chunks of traffic data could be insufficient to capture long-term attacks in a complex network, e.g., scan attacks. As a future work, the introduction of temporal information will be explored, thus analyzing the evolution of network status. %, and the usage of a different set of features for computing the representation will be investigated. 
\bibliographystyle{IEEEbib}
\bibliography{biblio}

\end{document}